# Representing Heuristic Knowledge in D-S Theory

Weiru Liu
Dept. of Artificial Intelligence
Univ. of Edinburgh
Edinburgh EH1 1HN

John G. Hughes    Michael F. McTear
Dept. of Information Systems
Univ. of Ulster at Jordanstown
Co. Antrim BT37 0QB,U.K.

## Abstract

The Dempster-Shafer theory of evidence has been used intensively to deal with uncertainty in knowledge-based systems. However the representation of uncertain relationships between evidence and hypothesis groups (heuristic knowledge) is still a major research problem. This paper presents an approach to representing such heuristic knowledge by **evidential mappings** which are defined on the basis of mass functions. The relationships between evidential mappings and multivalued mappings, as well as between evidential mappings and Bayesian multi- valued causal link models in Bayesian theory are discussed. Following this the detailed procedures for constructing evidential mappings for any set of heuristic rules are introduced. Several situations of belief propagation are discussed.

## 1 INTRODUCTION

In the design and implementation of expert systems and decision making systems, the problem of uncertain knowledge and evidence has to be solved. Several approaches can be used to deal with this problem, such as Mycin's certainty factors, Prospector's inference nets, fuzzy sets, Bayesian nets and Dempster-Shafer's belief functions. Generally speaking, there are two kinds of problem involving uncertainty: one is caused by uncertain evidence; another is caused by uncertain knowledge, i.e. heuristic knowledge. The former is a result of ill-defined concepts in the observation, or due to inaccuracy and poor reliability of the instruments used to make the observations. The latter is a result of weak implication which occurs when the expert or model builder is unable to establish a strong correlation between premise (or evidence) and conclusion (or hypotheses) (Bonissone and Tong 1985).

The Dempster-Shafer theory of evidence provides a flexible approach to representing uncertain evidence. This theory, which is claimed as an generalization of Bayesian inference (Shafer 1976, 1981), has the advantages of representing ignorance of evidence and narrowing the hypothesis space as a result of evidence accumulation. Several AI implementations have been undertaken (Laskey et al 1989, Lowrance et al 1986, Strat 1987, Wesley 1988, Yen 1989, Zarley et al 1988) based on the theory or extended versions of the theory (Laskey and Lehner 1989; Yen 1989). In this paper we argue that it is difficult to represent uncertain heuristic knowledge in this theory; however in most complex domains, heuristic knowledge plays an important role in solving problems.

Consider the following piece of heuristic knowledge: if $X$ is $X_1$, then $Y$ is $Y_1$ with a degree of belief $r_1$. If we get a piece of evidence which says that $X$ is $X_1$ with a degree of $a_1$, by invoking this rule we should be able to obtain the corresponding degree $y_1$ for $Y$ is $Y_1$. Certainly the value of $y_1$ must be a function $F$ of $a_1$ and $r_1$ (i.e. $y_1 = F(a_1, r_1)$). More generally, we suppose that a set of heuristic rules R includes:

$$R_1: \text{if } E_1 \text{ then } H_{11} \text{ with a degree of belief } r_{11};$$
$$H_{12} \text{ with a degree of belief } r_{12};$$
$$\ldots$$
$$R_2: \text{if } E_2 \text{ then } H_{21} \text{ with a degree of belief } r_{21};$$
$$H_{22} \text{ with a degree of belief } r_{22};$$
$$\ldots$$
$$\ldots \quad (1)$$

where $E_1, E_2, ..., E_n$ are values (or propositions) of the variable $E$, and $E_i$ is called an antecedent of rule $R_i$. A $H_{ij}$ in rule $R_i$ is a subset of the values (or propositions) of the variable H and it is called one of the conclusions of rule $R_i$. A $r_{ij}$ is called a rule strength.

Assume we have a piece of evidence which says that $E_1$ is confirmed with $a_1$, $E_2$ is confirmed with $a_2$, ..., $E_n$ is confirmed with $a_n$, how can we solve the following problems:

- what conditions should $\sum_i a_i$ satisfy?
- what conditions should $\sum_j r_{ij}$ satisfy?
- what is the function $F$ to determine $h_{ij}$(the degree

of belief on $H_{ij}$) from those $a_i$ and $r_{ij}$?

- if more than one set of rules is invoked and the same conclusion $H_{ij}$ is obtained, what will be the final degree of belief on $H_{ij}$ from those $h_{ij}, ..., h_{kl}$?

Generally, if the variable $E$ is a Cartesian product of variables $A, B, ..., C$, that is each $E_i$ is in a form of $(A_i \text{ and } B_j \text{ and} ... \text{and} C_k)$, assuming we know the evidence for $A, B, ..., C$, then

- what is the function $F'$ to determine the degree of belief on the premise $(A_i \text{ and } B_j \text{ and} ... \text{and} C_k)$?

These problems have been modelled in fuzzy theory using a fuzzy extension of modal logic, based on Zadeh's concepts of necessity and possibility (Prade 1981). They were also solved in Mycin's certainty factor model (Shortliffe and Buchanan 1976). Can these problems be solved in Dempster-Shafer theory?

In this paper we analyze these problems and propose our approaches for solving them by extending the theory of evidence. The paper is organized as follows. In section 2 the basics of Dempster- Shafer theory are introduced and the approach for representing heuristic knowledge by evidential mappings is described in which a matrix is used to represent the uncertain relationships between evidence and conclusions. In section 3 the relations between Bayesian inference and evidential mappings are examined in which it is proved that the multi-valued causal links between hypotheses space H and evidence space E (Pearl 1988) in Bayesian theory is consistent with the special case of evidential mappings. In section 4 the method of constructing a complete evidential mapping matrix for an evidential mapping of a heuristic rule is discussed. In section 5 belief propagation approaches are discussed for different situations. Finally a conclusion is given along with some consideration of related work.

## 2 REPRESENTING HEURISTIC KNOWLEDGE IN D-S THEORY

The Dempster-Shafer theory of evidence (which is also called the theory of belief functions (Smets 1988, Shafer 1990)) provides an alternative approach to drawing plausible conclusions from uncertain and incomplete evidence. It is a generalization of the Bayesian theory of subjective probability, it is more flexible, and it allows us to derive degrees of belief for a question from probabilities of a related question.

### 2.1 THE BASICS OF D-S THEORY

Suppose $\Theta$ is a finite set, which consists of mutually exclusive and exhaustive propositions of a problem or all values of a variable, $2^\Theta$ is the set of all subsets of $\Theta$. A function $Bel: 2^\Theta \to [0, 1]$ is called a belief function in Shafer (1976), if it satisfies the following conditions:

1. $Bel(\emptyset) = 0$;   2. $Bel(\Theta) = 1$;

3. for every positive integer $n$ and every collection $A_1, ..., A_n$ of subsets of $\Theta$,

$$Bel(A_1 \cup ... \cup A_n) \geq \sum_i Bel(A_i) - \sum_{i<j} Bel(A_i \cap A_j) + -... + (-1)^{n+1} Bel(A_1 \cap ... \cap A_n)$$

Such a set $\Theta$ is called a **frame of discernment**. By knowing a belief function on a frame of discernment, another function $m$ can be calculated as:

$$m(A) = \sum_{B \subseteq A} (-1)^{|A-B|} Bel(B) \quad \forall A \subseteq \Theta$$

where $|A - B|$ denotes the number of elements in the set of $A - B$.

The function $m$ is called a **basic probability assignment** *(bpa)* or a **mass function**. Obviously a mass function has the features that $m(\emptyset) = 0$ and $\sum m(A) = 1$ for all subsets $A$ of $\Theta$. A subset $A$ is called a focal element of the belief function $Bel$ if $m(A) > 0$. Recovering the belief function $Bel$ from a mass function $m$ is carried out by $Bel(B) = \sum_{A \subseteq B} m(A)$.

If all the focal elements of a belief function are singletons of $\Theta$, then the corresponding mass function $m$ is a Bayesian subjective probability distribution.

A belief function (or a mass function) on a frame $\Theta$ can either be directly obtained from a piece of evidence or calculated from a probability measure $P$ on the related frame T by a *multivalued mapping* $\Gamma$ between T and $\Theta$ (Dempster 1967), that is a multivalued mapping $\Gamma$ assigns each element $t$ of T to a subset $A$ of $\Theta$. The impact of several belief functions (or mass functions) on the same frame of discernment is obtained by using Dempster's rule of combination which treats Bayesian conditioning probabilities as a special case (Shafer 1976). Dempster's rule of combining two belief functions $Bel_1$ and $Bel_2$ can be defined by a relatively simple rule in terms of the corresponding mass functions $m_1$ and $m_2$.

$$m(C) = \frac{\sum_{A \cap B = C} m_1(A) m_2(B)}{1 - \sum_{A \cap B = \emptyset} m_1(A) m_2(B)}$$

This rule requires that the combined belief functions (or their mass functions) are independent. This condition has been further enhanced as *DS-independent* in Voorbraak (1991).

### 2.2 REPRESENTING HEURISTIC KNOWLEDGE IN D-S THEORY

It is obvious that a heuristic rule like {if $X$ is $X_1$ then $Y$ is $Y_1$ with a degree of belief $r_1$} cannot be directly represented in D-S theory. Some work concerning this topic was carried out previously (Ginsberg 1984, Yen 1988, Liu 1986, Hau and Kashyap 1990). We propose that evidential mappings which are defined on the basis of mass functions can be used to represent the uncertain relationships between evidence and conclusions.

**Definition 1** *An evidential mapping is the mapping from one frame of discernment to another, which represents causal links among elements of two frames of discernment in the form of mass functions. Formally an evidential mapping from frame $\Theta_E$ to frame $\Theta_H$ is a function $\Gamma^*: \Theta_E \longrightarrow 2^{2^{\Theta_H} \times [0,1]}$. The image of each element in $\Theta_E$, denoted by $\Gamma^*(e_i)$, is a collection of subset-mass pairs:*
$\Gamma^*(e_i) = \{(H_{i1}, f(e_i \to H_{i1})), ..., (H_{im}, f(e_i \to H_{im}))\}$
*and let $\Theta_i = \cup_{j=1}^m H_{ij}$ ($H_{ij} \subseteq \Theta_H$)*
*that satisfies the following conditions:*

a. $H_{ij} \neq \emptyset$      $j = 1, ..., m$
b. $f(e_i \to H_{ij}) > 0$      $j = 1, ..., m$
c. $\sum_j (e_i \to H_{ij}) = 1$

There is a set of heuristic rules, denoted as **R**, related to an evidential mapping, each of which is in the form of $R_i$:

$e_i \longrightarrow H_{i1} \; (f(e_i \to H_{i1})); \; ...; \; e_i \longrightarrow H_{im} \; (f(e_i \to H_{im}))$.

A rule states that if $e_i$ is true then the truth of the problem carried by $\Theta_H$ is in $H_{i1}$ with the degree of belief $f(e_i \to H_{i1})$ exactly committed to $H_{i1}$, ..., in $H_{im}$ with the degree of belief $f(e_i \to H_{im})$ exactly committed to $H_{im}$. The $e_i$ is called the antecedent of rule $R_i$ and it is an element of $\Theta_E$. $H_{ij}$ is called one of the conclusions of rule $R_i$, and it is a subset of $\Theta_H$. We name $\Theta_E$ and $\Theta_H$ as antecedent frame and conclusion frame of $R$ respectively. $f(e_i \to H_{ij})$ represents our belief exactly on $H_{ij}$ given condition $e_i$, and it is in the range of $[0,1]$.

The corresponding matrix of this evidential mapping is:

| row/col | $H_1$ | $H_2$ | ... | $\Theta$ | |
|---|---|---|---|---|---|
| $\{e_1\}$ | $m_{11}$ | $m_{12}$ | ... | $m_{1l}$ | |
| $\{e_2\}$ | $m_{21}$ | $m_{22}$ | ... | $m_{2l}$ | $= M$ |
| ... | | | | | |
| $\{e_n\}$ | $m_{n1}$ | $m_{n2}$ | ... | $m_{nl}$ | |

The size of matrix $M$ is $n \times l$ where $n$ is the number of elements in $\Theta_E$ and $l$ equals $|2^{\Theta_H}| - 1$ (except $\emptyset$). $H_i$ is a subset of the elements of $\Theta_H$. For any $H_{ij}$ appearing in $(H_{ij}, f(e_i \to H_{ij}))$ there is $H_k$ where $H_k = H_{ij}$. The $(i,k)$-th entry of $M$ is defined as $m_{ik}$ which equals $f(e_i \to H_{ij})$ if the pair $(H_{ij}, f(e_i \to H_{ij}))$ is an element of $\Gamma^*(e_i)$ and $H_k = H_{ij}$; otherwise $m_{ik}$ equals 0. Thus those $m_{i1}, m_{i2}, ..., m_{il}$ of line $i$ must satisfy the condition $\sum_j m_{ij} = 1$. More precisely based on $m_{i1}, m_{i2}, ..., m_{il}$, we define a function $m_i$. In fact $m_i$ is a mass function on $\Theta_E \times \Theta_H$, with its focal elements as $A_{i1} = \{(x,y) | x \in \neg\{e_i\} \text{ or } y \in H_{i1}\}, ..., A_{im} = \{(x,y) | x \in \neg\{e_i\} \text{ or } y \in H_{im}\}$ and $m_i(A_{ij}) = m_{ij}$ for $j = 1, ..., l$. So there are in total $n$ mass functions on frame $\Theta_E \times \Theta_H$. But we define that the combination of any two of the above mass functions is meaningless.

In order to identify each row and column in $M$ we call $H_k$ the *title of column $k$*, $\{e_i\}$ the *title of row $i$*. We also call $[\{e_1\}, \{e_2\}, ..., \{e_n\}]$ and $[H_1, H_2, ..., \Theta]$ the *row title vector* and *column title vector* of $M$ respectively. When we mention a matrix $M$ of an evidential mapping, we assume the row title vector and the column title vector are known. Thus for any given evidential mapping the related heuristic rule set and the matrix are unique.

An evidential mapping from $\Theta_E$ to $\Theta_H$ states that for two related questions represented by $\Theta_E$ and $\Theta_H$, if the truth for the question represented by $\Theta_E$ is $e_i$ then the truth for the question represented by $\Theta_H$ is in a set $\Theta_i$, but $e_i$ has different inter relationships with different subsets of $\Theta_i$. The $f(e_i \to H_{ij})$ is used to reflect the sensitivity or strength of interrelation between $e_i$ and $H_{ij}$. Certainly the total strength should be 1.

**Example 1:** If an evidential mapping $\Gamma^*$ specifies mapping from an evidence space $\Theta_E$ to a hypothesis space $\Theta_H$ as:

$\Gamma^*(e_1) = \{(\{a_1, a_2\}, 0.7), (\{a_3, a_4\}, 0.3)\}$
$\Gamma^*(e_2) = \{(\{a_2, a_3\}, 0.8), (\Theta_H, 0.2)\}$
$\Gamma^*(e_3) = \{(\{a_4, a_5\}, 0.9), (\Theta_H, 0.1)\}$

and a related set of heuristic rules is

R:    $e_1 \longrightarrow \{a_1, a_2\}$ (0.7);    $e_1 \longrightarrow \{a_3, a_4\}$ (0.3).
     $e_2 \longrightarrow \{a_2, a_3\}$ (0.8);    $e_2 \longrightarrow \Theta_H$ (0.2).
     $e_3 \longrightarrow \{a_4, a_5\}$ (0.9);    $e_3 \longrightarrow \Theta_H$ (0.1).

where $\Theta_E = \{e_1, e_2, e_3\}$ and $\Theta_H = \{a_1, a_2, a_3, a_4, a_5\}$, then the matrix $M$ has $2^5 - 1$ columns, most of which have only zero $m_{ij}$ such as columns $\{a_1\}, \{a_1, a_2, a_3\}$. Usually a matrix becomes too big when $\Theta_H$ contains several elements. So we delete all those columns which have only zero $m_{ij}$ and form another matrix. We call such a simplified matrix the **Basic Matrix** and denote it as **BM**. Thus the title vector of a basic matrix of an evidential mapping only contains those $H_{ij}$ which appear in $\Gamma^*(e_i)$. The BM of this evidential mapping in the above example is

| row/col | $\{a_1, a_2\}$ | $\{a_2, a_3\}$ | $\{a_3, a_4\}$ | $\{a_4, a_5\}$ | $\Theta_H$ |
|---|---|---|---|---|---|
| $\{e_1\}$ | 0.7 | 0.0 | 0.3 | 0.0 | 0.0 |
| $\{e_2\}$ | 0.0 | 0.8 | 0.0 | 0.0 | 0.2 |
| $\{e_3\}$ | 0.0 | 0.0 | 0.0 | 0.9 | 0.1 |

with row title vector $[\{e_1\}, \{e_2\}, \{e_3\}]$ and column title vector $[\{a_1, a_2\}, \{a_2, a_3\}, \{a_3, a_4\}, \{a_4, a_5\}, \Theta_H]$.

Obviously, multivalued mappings in section 2.1 and Bayesian multi-valued causal link models (Pearl 1988) can all be represented using such evidential mappings.

**Corollary 1** *If all the $m_{ij}$ in a basic matrix BM of an evidential mapping from $\Theta_E$ to $\Theta_H$ are either 1 or 0 then the evidential mapping is a multivalued mapping. For any $e_i$, the mass function $m_i$ on $\Theta_E \times \Theta_H$ is a simple support function with a focal element $A_{ij} (A_{ij} = \{(x,y) | x \in \neg\{e_i\} \text{ or } y \in H_{ij}\})$, and $m_i(A_{ij}) = 1$.*

**Corollary 2** *If a basic matrix BM has $|\Theta_H|$ columns, and the titles of all columns are singletons of $\Theta_H$ then*

the evidential mapping from $\Theta_E$ to $\Theta_H$ of this matrix is exactly a Bayesian multi-valued causal link model. For any $e_i$, the mass function $m_i$ on $\Theta_E \times \Theta_H$ is a Bayesian probability distribution. We refer to this kind of evidential mappings as Bayesian evidential mappings.

If a piece of evidence gives a probability distribution $P$ on $\Theta_E$, then a new function $m$ on $\Theta_H$ can be calculated by the evidential mapping from $\Theta_E$ to $\Theta_H$:

$m(H_k) = \sum_i P(e_i) \times m_{ik} = \sum_i P(e_i) \times f(e_i \rightarrow H_k)$
(When $H_k$ is the title of a column)        (2)
$m(H_k) = 0$         Otherwise

The function $m$ is a basic probability assignment in the hypothesis space and has the following features:
1) $m(\emptyset) = 0$ and 2) $\sum_k m(H_k) = 1$ where $H_k \subseteq \Theta$.

This can be proved by the following according to definition 1, probability distribution $P$ and features of a mass function.

$\sum_k m(H_k) = \sum_k \sum_i P(e_i) \times f(e_i \rightarrow H_k)$
$= \sum_i \sum_k P(e_i) \times f(e_i \rightarrow H_k)$
$= (\sum_i P(e_i))(\sum_k f(e_i \rightarrow H_k))$
$= (\sum_i P(e_i))(\sum_j f(e_i \rightarrow H_{ij})) = 1 \times 1 = 1$
(because there exists an $H_{ij}$ such that $H_k = H_{ij}$)

**Corollary 3** *A function $m$ is a mass function on frame $\Theta_H$ if it is given by formula 2 under the condition that $P$ is a probability distribution on space $\Theta_E$ and $\Gamma^*$ is an evidential mapping from $\Theta_E$ to $\Theta_H$.*

The theoretical support of the formula (2) is Bayes' formula
$$P(A) = \sum_i P(A|B_i)P(B_i)$$
where $B_i$ is an element of an exhaustive and mutually exclusive event set (Pearl 1988).

We suppose that any evidence $e$ giving $P(B_i/e)$ has no effect on $P(A|B_i)$. This rule is also called Jeffrey's rule of conditioning (Jeffrey 1965, Shafer 1981).

### 2.3  CREATING EVIDENTIAL MAPPINGS FOR INCOMPLETE HEURISTIC RULE SETS

We have seen in the above section that an evidential mapping can be associated with a set of heuristic rules. The other way around, given a set of heuristic rules in the form of (1), if all the antecedents of rules can form a frame of discernment $\Theta_E$, all the conclusions of rules can form another frame of discernment $\Theta_H$, and for any heuristic rule $R_i$ the sum of $r_{ij}$ (for $j = 1, .., m$) is 1, then an evidential mapping can be established between $\Theta_E$ and $\Theta_H$. Unfortunately, the antecedents (or conclusions) of a set of rules normally cannot form a frame of discernment which is mutually exclusive and exhaustive and usually the sum of $r_{ij}$ for rule $R_i$ is less than 1. For example if there is only one rule in a set of heuristic rules: if $X$ is $X_1$ then $Y$ is $Y_1$ with a degree of belief $r_1$, then the antecedent ($X$ is $X_1$) itself does not form a frame of discernment at all nor does the conclusion ($Y$ is $Y_1$).

**Definition 2** *If at least one of the antecedent frame and conclusion frame of a heuristic rule set $R$ is not a frame of discernment or there is a rule $R_i$ in rule set $R$ where $\sum_j r_{ij} < 1$, then we define such a heuristic rule set as an* **incomplete heuristic rule set**. *Otherwise we call it a complete heuristic rule set.*

**Corollary 4** *Given an incomplete heuristic rule set $R$, let $E = \{e_1, ..., e_n\}$ represent the antecedent set, and $H = \{h_1, ..., h_m\}$ represent the conclusion set of $R$,*

- *if $E$ is not a frame of discernment then define $e_{n+1} = \neg(e_1 \vee ... \vee e_n)$ and $\Theta_E = \{e_{n+1}\} \cup E$; otherwise define $\Theta_E = E$.*

- *if $H$ is not a frame of discernment then define $h_{m+1} = \neg(h_1 \vee ... \vee h_m)$ and $\Theta_H = \{h_{m+1}\} \cup H$; otherwise define $\Theta_H = H$.*

- *if $e_{n+1}$ exists then add the rule $R_{n+1} : (e_{n+1} \longrightarrow \Theta_{H \ (1)})$ to rule set $R$.*

*Then $\Theta_E$ and $\Theta_H$ are two frames of discernment representing the antecedent frame and the conclusion frame of $R$ respectively.*

**Corollary 5** *For each rule $R_i$ in $R$, if $\sum_j r_{ij} < 1$ then we add an extra conclusion $\Theta_H$ with belief $r_{ih} = (1 - \sum_j r_{ij})$ to $R_i$. That is, if the original $R_i$ is*

$R_i : \quad e_i \longrightarrow H_{i1 \ (r_{i1})}; ...; e_i \longrightarrow H_{im \ (r_{im})}.$
*then a new $R'_i$ is*
$R'_i : \quad e_i \longrightarrow H_{i1 \ (r_{i1})}; ...; e_i \longrightarrow H_{im \ (r_{im})};$
$e_i \longrightarrow \Theta_{H \ (r_{ih})}$ *where $r_{ih} = 1 - \sum_j r_{ij}$.*

Now the heuristic rule set $R$ is complete and an evidential mapping from $\Theta_E$ to $\Theta_H$ can be created. In fact the added part of a rule represents our ignorance. In other words, based on the current knowledge of a specific domain, we have no knowledge to identify any more ad-hoc relationships among elements of reasons and results.

**Example 2:** Suppose we have a rule set $R$ which consists of a rule as follows:

Smoke alarm is ringing $\longrightarrow$ There is a fire $_{(0.9)}$.

Constructing $\Theta_E = \{($ smoke alarm is ringing), not(smoke alarm is ringing)$\}$, $\Theta_H = \{($ there is a fire), not(there is a fire)$\}$ based on corollary 4, and a new rule set $R'$ based on corollary 5 and $R'$ has:

$R_1$: Smoke alarm is ringing $\longrightarrow$ There is a fire $_{(0.9)}$;
Smoke alarm is ringing $\longrightarrow \Theta_{H \ (0.1)}$.
$R_2$: Not (Smoke alarm is ringing) $\longrightarrow \Theta_{H \ (1)}$.

This rule set can be associated with an evidential mapping from $\Theta_E$ to $\Theta_H$. In particular, if $\Theta_H$ is the same as $\Theta_E$ then the corresponding evidential mapping represents self-relations of $\Theta_E$ (it is also called delta-$\Theta_E$ compatibility relation by Lowrance et al (1986)). Now we can represent any heuristic rule set (either complete or incomplete) in the Dempster-Shafer theory of evidence by the means of evidential mappings. In the following we simply use a triple $(R, \Theta_E, \Theta_H)$ to represent an evidential mapping where $R$ is a heuristic rule set, $\Theta_E$ is the antecedent frame of discernment of $R$ and $\Theta_H$ is the conclusion frame of discernment of $R$.

## 3  THE RELATION BETWEEN EVIDENTIAL MAPPINGS AND BAYESIAN CONDITIONAL PROBABILITIES

The Dempster-Shafer theory of evidence as an generalization of Bayesian inference includes two meanings: mass functions are the general form of Bayesian subjective probabilities in representing evidence; Bayesian conditional probabilities are a special case of Dempster's rule of combination (Shafer 1976). Pearl (1988) gave general formula to calculate posterior-probabilities (on hypotheses) or predict future events in multi-valued causal link models of Bayesian theory when a set of evidence (for evidence variable) is given. In fact, Pearl's work is the extension of traditional Bayesian inference theory to the situation when the relationships among elements of an evidence space and a hypothesis space are multi-valued causal mappings. In this section we prove that Bayesian inference performed on multi-valued causal link models can be carried out in D-S theory by using evidential mappings.

### 3.1  PREDICTING FUTURE EVENTS IN D-S THEORY

**Example 3:** Let $S$ be a variable for "alarm sound" and $D$ for "a person's call". If we use the same capital letter to represent both a variable name and the name of the frame which includes all the values of the variable, we have $S = \{alarm\ on, alarm\ off\}$ and $D = \{a\ person\ will\ call,\ a\ person\ will\ not\ call\}$ each of which represents an exhaustive and mutually exclusive set of propositions. Suppose the causal link between $S$ and $D$ is

| $S/D$ | will call | will notcall |
|---|---|---|
| alarm on | 0.7 | 0.3 |
| alarm off | 0.0 | 1.0 |

Bayesian inference produces

$$P(d_i) = \sum_j P(d_i|s_j)P(s_j) \quad (3)$$

which is a shorthand notation for the statement

$$P(d_i \mid e) = \sum_j P(d_i|s_j,e)P(s_j \mid e)$$

where $d_i$ is an element of $D$ and $s_j$ is an element of $S$ and we assume that a piece of evidence has no effect on the causal link between $S$ and $D$. Given a probability distribution of a piece of evidence on $S$, the probabilities on $D$ can be calculated from formula (3).

Suppose $P(s_1 = on) = 0.2686$, $P(s_2 = off) = 0.7314$, then $P(d_1 = will\ call) = \sum_j P(d_1 \mid s_j)P(s_j)$

$$= [0.2686, 0.7314] \times \begin{bmatrix} 0.7 \\ 0.0 \end{bmatrix} = 0.188 \quad (4).$$

This is called *predict future events* by Pearl in Bayesian inference. Obviously the causal link above forms an evidential mapping from $S$ to $D$ in Dempster-Shafer theory. In the condition of prior probabilities $P(s_1 = on) = 0.2682$, $P(s_2 = off) = 0.7314$, applying formula (2) we get a mass function on $D$ which is the same as that showed in formula (4).
$m(d_1) = \sum_i P(s_i) \times f(s_i \rightarrow \{d_1\})$

$$= 0.2686 \times 0.7 + 0.7314 \times 0.0 = 0.188$$

$m(d_2) = \sum_i P(s_i) \times f(s_i \rightarrow \{d_2\})$

$$= 0.2686 \times 0.3 + 0.7314 \times 1.0 = 0.812$$

In Bayesian multi-valued causal link models, the causal link between the hypothesis space $H$ and the evidence space $E$ is identified by a $n \times m$ matrix $M$, where $n$ and $m$ are the numbers of values of $H$ and $E$ respectively, and the $(i,j)$-th entry of $M$ is $M_{ij} = P(e_j \mid h_i)$ (Pearl 1988).

It is easy to see (corollary 3) that the causal link model above is consistent with the special case of evidential mappings. The mass function on $D$ obtained from formula (2) is exactly the same as the probability distribution on $D$ obtained in Bayesian inference.

### 3.2  CALCULATING POSTERIOR PROBABILITIES IN D-S THEORY

Furthermore, in Bayesian multi-valued causal link models, given a prior probability distribution on hypothesis space $H$, causal link matrix $M$ with $M_{ij} = P(e_j \mid h_i)$

|  | $\{e_1\}$ | $\{e_2\}$ | ... | $\{e_m\}$ |  |
|---|---|---|---|---|---|
| $\{h_1\}$ | $p(e_1 \mid h_1)$ | $p(e_2 \mid h_1)$ | ... | $p(e_m \mid h_1)$ |  |
| $\{h_2\}$ | $p(e_1 \mid h_2)$ | $p(e_2 \mid h_2)$ | ... | $p(e_m \mid h_2)$ | $= M$ |
|  | ... |  |  |  |  |
| $\{h_n\}$ | $p(e_1 \mid h_n)$ | $p(e_2 \mid h_n)$ | ... | $p(e_m \mid h_n)$ |  |

and a set of evidence $e^1, e^2, ..., e^N$ on evidence space $E$, then posterior-probability $P(h_i \mid e^1, e^2, ..., e^N)$ on $h_i$ of $H$ is:

$$P(h_i \mid e^1, e^2, ..., e^N) = \alpha P(e^1, e^2, ..., e^N \mid h_i)P(h_i) \quad (5)$$

where $\alpha = [P(e^1, e^2, ..., e^N)]^{-1}$ is a normalizing constant to be computed by requiring that Eq.(5) sum to unity. Assuming $e^1, e^2, ..., e^N$ are independent with each other and conditional independence of respect to each $h_i$, Pearl (1988) indicated that,

$$P(h_i \mid e^1, e^2, ..., e^N) = \alpha P(h_i)[\Pi_{k=1}^N P(e^k \mid h_i)] \quad (6)$$

Here we should make it clear that Pearl assumes that for each piece of evidence $e^k$ there exists an element $e_j$ in $E$ where $p(e_j) = 1$ given by $e^k$ so that $P(e^k \mid h_i) = P(e_j \mid h_i)$. Can these posterior probabilities be calculated in D-S theory using evidential mappings based on the above causal link matrix under these assumptions? The following theorem indicates that they can.

**Theorem 1** [1] *Let $E$ and $H$ be two frames of discernment, $\Gamma^*$ be a Bayesian evidential mapping from $H$ to $E$, $BM$ be the basic matrix of the mapping $\Gamma^*$ with $(i,j)$-th entry as $p(e_j \mid h_i)$. Assume the prior probability on $h_i$ of $H$ is $p(h_i)$, a set of evidence on $E$ is $e^1, e^2, ..., e^N$ for each of which there exists an $e_l$ where $p(e_l) = 1$. Then the final belief function Bel on $H$ using D-S theory is*

$$Bel(h_i) = \alpha p(h_i)[\Pi_{k=1}^N p(e^k \mid h_i)] \qquad (7)$$

$$where \quad \alpha = (\sum_1^n (p(h_i)[\Pi_{k=1}^N p(e^k \mid h_i)]))^{-1}$$

*and $p(e^k \mid h_i) = p(e_l \mid h_i)$ for each $k$ when the evidence $e^k$ makes $p(e_l) = 1$.*

## 4 CONSTRUCTING COMPLETE EVIDENTIAL MAPPING MATRICES TO PROPAGATE MASS FUNCTIONS FROM AN EVIDENCE SPACE $\Theta_E$ TO A HYPOTHESIS SPACE $\Theta_H$

In Dempster-Shafer theory a miltivalued mapping is used to calculate a mass function on a frame based on either a probability distribution or a mass function on another frame (Lowrance et al 1986, Zarley 1988, Laskey et al 1989). What we have assumed in the previous two sections is that a piece of evidence on an evidence space (a frame of discernment $\Theta_E$) is represented in the form of Bayesian subjective probabilities. A mass function on $\Theta_H$ will be obtained based on the probability distribution on $\Theta_E$ through an evidential mapping from $\Theta_E$ to $\Theta_H$.

In section 2, we gave the definition of evidential mappings. Let $\Theta_E$ and $\Theta_H$ be two frames of discernment, $\Gamma^*$ be an evidential mapping from $\Theta_E$ to $\Theta_H$. Assuming a piece of evidence indicates that $m(E) = p, m(\Theta_E) = 1 - p$, $E$ is a subset of $\Theta_E$, what is the impact of the evidence on $\Theta_H$? Obviously the impact of the evidence on $\Theta_H$ is easy to be got if $\Gamma^*$ is a multivalued mapping. But it is not so easy when $\Gamma^*$ is an evidential mapping.

[1] The proof of this theorem and other theorem, examples, and further discussion about evidential mappings and related work are given in the full version of the paper which is available on request (Liu et al 1992).

In this section, we introduce the approach to constructing complete evidential mapping matrices between two frames. A complete evidential mapping matrix between two frames allows the propagation of a mass function from one frame to another.

**Definition 3** *If $\Gamma^*$ is an evidential mapping from $\Theta_E$ to $\Theta_H$, $BM$ is the basic matrix of $\Gamma^*$ with $m_{ij}$ as its $(i,j)$-th entry, the titles of rows of $BM$ are $\{e_1\}$, ..., $\{e_n\} \subseteq \Theta_E$, the titles of columns of $BM$ are $A_1$, ..., $A_m \subseteq \Theta_H$, then a matrix is called a complete evidential mapping matrix of $BM$, denoted as $CEM$, if it is defined as:*

1. *all the subsets of $\Theta_E$ except $\emptyset$ are titles of rows of $CEM$ and $\{e_1\}$, ..., $\{e_n\}$ are the first $n$ row titles; $A_1$, ..., $A_m$ are the titles of the first $m$ columns of $CEM$.*

2. *the $m_{ij}$ of $BM$ is the value of $(i,j)$-th entry of $CEM$ and denoted as $m'_{ij}$.*

3. *for a row $l$ with the title $E$, and $l > n$, suppose $E = \{e_{l_1}, e_{l_2} ..., e_{l_k}\}$, then $(l,j)$-th entry of $CEM$ is*

$m_{lj} = (m_{l_1j} + m_{l_2j} + ... + m_{l_kj})/k \quad$ *if all $m_{l_ij} \neq 0$*
*for $i = 1, ...k$  $m_{lj} = 0 \quad$ otherwise*

4. *for $m_{lj} = 0$, create a column $r$ with the title $A_r$*

*let $A_r = \cup_i \Theta_i \quad$ for $i = l_1, ..., l_k$. (for $\Theta_i$ see definition 1 in 2.2)*

*if $A_r$ is not an element of column title vector, then add $A_r$ as an column title and the value of $(l,r)$-th entry is $m_{lr} = (m_{l_1j} + m_{l_2j} + ... + m_{l_kj})/k$. Otherwise there is a column $r'$ satisfying $A_{r'} = A_r$, we update $m_{lr'}$ as $m_{lr'} + (m_{l_1j} + m_{l_2j} + ... + m_{l_kj})/k$*

5. *for any other entry $(x,y)$, define $m_{xy} = 0$.*

Obviously we have the unequal formula
$$max(m_{l_1j}, m_{l_2j}..., m_{l_kj}) \geq (m_{l_1j} + m_{l_2j} + ... + m_{l_kj})/k$$
$$\geq min(m_{l_1j}, m_{l_2j}, ..., m_{l_kj}) \qquad (8)$$

The basic idea of constructing a complete evidential mapping matrix is that if the causal links from $e_{l1}, e_{l2} ..., e_{lk}$ to $A'$ are $m_{l_1j}, m_{l_2j} ..., m_{l_kj}$ respectively, then the causal link from $\{e_{l1}, e_{l2}, ..., e_{lk}\}$ to $A'$ is something between $max(m_{l_1j}, m_{l_2j}, ..., m_{l_kj})$ and $min(m_{l_1j}, m_{l_2j}, ..., m_{l_kj})$. Here we use the average value of $m_{l_1j}, m_{l_2j}, ..., m_{l_kj}$ to represent approximate causal link from $\{e_{l1}, e_{l2} ..., e_{lk}\}$ to $A'$.

It is easy to prove that a CEM is a basic matrix of an evidential mapping from $2^{\Theta_E}$ to $\Theta_H$. So any piece of evidence, which is in the form of $bpa$ on $\Theta_E$ can be propagated to $\Theta_H$ through the CEM. If a BM is the matrix of an multivalued mapping then its related CEM is also associated with the same multivalued mapping.

Certainly if a rule in a rule set specifies the causal link between a subset $E$ of $\Theta_E$ and $A_1$, ..., $A_n$ of $\Theta_H$, then the values of row $i$, with the row title as $E$, are $(f(E \to A_1), ..., f(E \to A_n))$ in CEM. But these $f(E \to A_i)$ must satisfy the condition of (8).

## 5 PROPAGATING BELIEFS USING HEURISTIC KNOWLEDGE

Belief propagation in a rule based system as described above indicates that, given belief functions on an antecedent frame and a set of rules with rule strengths in the form of mass functions, the belief functions on the conclusion frame can be deduced. If $(R, \Theta_E, \Theta_H)$, $(R', \Theta_H, \Theta'_H)$, $(R'', \Theta'_E, \Theta_H)$, $(R_1, \Theta_E, \Theta_H)$ and $(R_2, \Theta_E, \Theta_H)$ are five triples associated with five evidential mappings, generally we need to solve the following belief propagation problems:

(i). given a piece of evidence on $\Theta_E$, $(R, \Theta_E, \Theta_H)$ is known, to deduce belief on $\Theta_H$.

(ii). given a piece of evidence on $\Theta_E$, $(R, \Theta_E, \Theta_H)$, and $(R', \Theta_H, \Theta'_H)$ are known, to deduce belief on $\Theta'_H$.

(iii). given two pieces of evidence on $\Theta_E$ and $\Theta'_E$ respectively, $(R, \Theta_E, \Theta_H)$ and $(R'', \Theta'_E, \Theta_H)$ are known, to deduce belief on $\Theta_H$.

(iv). given a piece of evidence on $\Theta_E$, $(R_1, \Theta_E, \Theta_H)$ and $(R_2, \Theta_E, \Theta_H)$ are given independently, to deduce belief on $\Theta_H$.

(v). given several pieces of evidence each of which is on $A, B, ..., C$ respectively, $(R, \Theta_E, \Theta_H)$ is known where $\Theta_E = A \times B \times ... \times C$, to deduce belief on $\Theta_H$.

These problems can be solved by the following three theorems.

**Theorem 2** *Let $(R, \Theta_E, \Theta_H)$ be a triple associated with an evidential mapping $\Gamma^*$, BM and CEM are the basic matrix and the complete evidential mapping matrix of $\Gamma^*$, if a mass function $m$ on $\Theta_E$ is known, then a mass function $m_1$ on $\Theta_H$ is calculated by the formula*

$$[m_1(H_1)...m_1(H_m)] = [m(E_1)...m(E_n)] \times CEM \quad (9)$$

*where $[E_1, ..., E_n]$ is the row title vector of CEM, and $[H_1, ..., H_m]$ is the column title vector of CEM.*

Specifically, if $m$ is a Bayesian subjective probability assignment then $m_1$ on $\Theta_H$ is calculated by

$$[m_1(H_1)...m_1(H_m)] = [m(e_1)...m(e_n)] \times CEM$$

**Theorem 3** *Let $(R, \Theta_E, \Theta_H)$ and $(R', \Theta_H, \Theta'_H)$ be two triples associated with two evidential mappings $\Gamma^*$ and $\Gamma'^*$, $CEM_1$ and $CEM_2$ are two complete evidential mapping matrices of $\Gamma^*$ and $\Gamma'^*$, if a mass function $m$ on $\Theta_E$ is known, then a mass function $m_1$ on $\Theta'_H$ is calculated by the formula*

$$[m_1(H'_1)...m_1(H'_m)] = [m(E_1)...m(E_n)] \times CEM$$

*where $CEM = CEM_1 \times CEM_2$.*

Theorem 3 indicates that if we know a series of evidential mappings from $\Theta_{E_1}$ to $\Theta_{E_2}$, ..., from $\Theta_{E_{n-1}}$ to $\Theta_{E_n}$ and those $CEM_i$ of evidential mappings from $\Theta_{E_i}$ to $\Theta_{E_{i+1}}$ for $i = 1,...,n-1$, then we will get an evidential mapping from $\Theta_{E_1}$ to $\Theta_{E_n}$ with its CEM as $CEM_1 \times ... \times CEM_{n-1}$.

Theorems 2 and 3 can be used to solve problems (i) and (ii). Dempster's combination rule is used to deal with the problem in (iii) where we suppose that any two pieces of evidence bearing on the same frame are DS-independent (Voorbraak1991). DS-independence will guarantee that if we use Dempster's rule to combine two probability distributions we should obtain the same result as what we get from Bayesian theory.

**Theorem 4** *Let $(R_1, \Theta_E, \Theta_H)$ and $(R_2, \Theta_E, \Theta_H)$ be two triples associated with two evidential mappings $\Gamma^*$ and $\Gamma'^*$, $m_i$ and $m'_i$ are two mass functions indicating causal links from $e_i$ to $\Theta_H$ in $\Gamma^*$ and $\Gamma'^*$ respectively (for $i = 1, ..., n$), that is*

$\Gamma^*(e_i) = \{(H_{i1}, f(e_i \to H_{i1})), ...,$
$(H_{in'}, f(e_i \to H_{in'}))\}$
$m_i(A_{il}) = f(e_i \to H_{il})$
*where $A_{il} = \{(x,y) | x \in \neg\{e_i\} \text{ or } y \in H_{il}\}$ for $l = 1, ..., n'$*
*and*
$\Gamma'^*(e_i) = \{(H'_{i1}, f(e_i \to H'_{i1})), ...,$
$(H'_{in''}, f(e_i \to H'_{in''}))\}$
$m'_i(A'_{ir}) = f(e_i \to H'_{ir})$
*where $A'_{ir} = \{(x,y) | x \in \neg\{e_i\} \text{ or } y \in H'_{ir}\}$ for $r = 1, ..., n''$*

*then the joint impact of two evidential mappings is the third evidential mapping $\Gamma''^*$ from $\Theta_E$ to $\Theta_H$ in which:*

$\Gamma''^*(e_i) = \{(H''_{i1}, f(e_i \to H''_{i1})), ...,$
$(H''_{ik}, f(e_i \to H''_{ik}))\}$ \quad (10)
$m_i(A''_{ij}) = f(e_i \to H''_{ij}) = m_i \oplus m'_i(A''_{ij})$ *for* $j = 1, ..., k$
*where* $A''_{ij} = A_{il} \cap A'_{ir}$, *and* $H''_{ij} = H_{il} \cap H'_{ir}$.

Here $\oplus$ indicates that Dempster's rule is used to combine $m_i$ and $m'_i$.

The meaning of this theorem is that if there are two independent heuristic rule sets (as in figure 1) given by different domain experts respectively, each of those specifies one kind of causal link from frame $\Theta_E$ to frame $\Theta_H$, then the joint impact of the two causal links can be substituted by the third heuristic rule set which is produced from them.

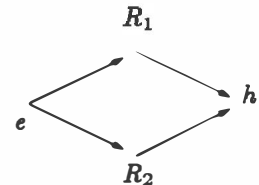

Figure 1: Two evidential mappings from $e$ to $h$.

Here we need to address the issue that the meaning of this theorem is different from using theorem 2 twice through two evidential mappings. Using theorem 2 in that way gives a wrong result because of the dependency of two mass functions on $\Theta_H$.

For problem (v), given a triple of an evidential mapping $(R, \Theta_E, \Theta_H)$, $\Theta_E = A \times B \times ... \times C$, and a series of evidence in the form of mass functions on $A, B, ..., C$ respectively, we must first get the joint mass function on $\Theta_E$ in order to obtain the impact of those pieces of evidence on $\Theta_H$. Shafer's partition theory and technique (Shafer 1976, Shafer, Shenoy and Mellouli 1987; Shafer and Logan 1987) provide a sound background for propagating mass functions (or belief functions) from $A, B, ..., C$ to their Cartesian product frame of discernment $\Theta_E$. Thus applying theorem 2 the mass function on $\Theta_H$ will be calculated. Certainly computational complexity is a major problem which has been widely researched (Barnett 1981, Shafer and Logan 1987).

Generally speaking, when $\Theta_E = A \times B \times ... \times C$, we can solve problem (v) by following steps:

1) establishing evidential mappings $\Gamma_A, \Gamma_B, ..., \Gamma_C$ (in fact they are multivalued mappings) from $A, B, ..., C$ to $\Theta_E$ respectively.

$$\Gamma_A(a_i) = \{(\{a_i\} \times B \times ... \times C, 1)\} \text{ for each } a_i \text{ in } A$$
$$\Gamma_B(b_i) = \{(A \times \{b_i\} \times ... \times C, 1)\} \text{ for each } b_i \text{ in } B$$
$$...\qquad(11)$$
$$\Gamma_C(c_i) = \{(A \times B \times ... \times \{c_i\}, 1)\} \text{ for each } c_i \text{ in } C$$

2) given a series of pieces of evidence on $A, B, ..., C$, based on those evidential mappings, geting a number of mass functions on the joint frame from each simple frame.

3) suppose $A, B, ..., C$ are different from each other and the pieces of evidence are independent, using Dempster's combination rule to get the final mass function on $\Theta_E$.

4) so based on this final mass function on $\Theta_E$ and an evidential mapping associated with $(R, \Theta_E, \Theta_H)$, applying theorem 2 eventually to deduce a mass function on $\Theta_H$.

## 6 CONCLUSION

### 6.1 RELATED WORK

Several approaches to dealing with heuristic knowledge in the Dempster-Shafer theory of evidence have been proposed (Ginsberg 1984, Liu 1986, Yen 1988, Hau and Kashyap 1990). Pearl also mentions this problem in Pearl (1990). The approach proposed in this paper is different from those approaches among which Liu's approach and Yen's approach are two proper subsets of our evidential mappings. In Ginsberg's as well as in Hau and Kashyap's representation formalisms of heuristic rules, a rule is associated with a pair of real numbers between [0,1] in the form of if $E$ then $H$ with uncertainty $[c, d]$.

The meaning of $c$ and $d$ defined by Ginsberg is: $c$ is the extent to which we believe a given proposition to be confirmed by the available evidence, and $d$ is the extent to which it is disconfirmed.

That is:     $E \to H$ with $c$ and $E \to \neg H$ with $d$.

While Hau and Kashyap gave two explanations:

1. $c$ is the credibility to which $E$ supports $H$, $d$ is the plausibility to which $E$ supports $H$, so $1 - d$ is the degree to which $E$ supports $\neg H$.

That is     $E \to H$ with $c$ and $E \to H$ with $1 - d$.

2. Let $\Theta = \Theta_E \oplus \Theta_H$ where $\Theta_E$ and $\Theta_H$ are frames of discernment of $E$ and $H$, then $c$ and $d$ form a mass function on $\Theta$

$$m(A) = c, m(\neg A) = 1 - d, \text{ and } m(\Theta) = d - c.$$
$$\text{where } A = \{(x, y) | x \in \neg E \text{ or } y \in H\}$$

Obviously Ginsberg's representation can be incorporated into an evidential mapping from $\Theta_E$ to $\Theta_H$ by the rule set $R$ below.

$$R: \quad E \to H_{(c)}; E \to \neg H_{(d)}; E \to \Theta_{H\,(1-d-c)}.$$
$$\neg E \to \Theta_{H\,(1)}.$$

Hau and Kashyap's first explanation can also be incorporated into an evidential mapping from $\Theta_E$ to $\Theta_H$ by the rule set $R'$

$$R': \quad E \to H_{(c)}; E \to \neg H_{(1-d)}; E \to \Theta_{H\,(d-c)}.$$
$$\neg E \to \Theta_{H\,(1)}.$$

In fact the second explanation given by Hau and Kashyap is to construct a mass function (furthermore a belief function) on a joint frame of discernment. Similar explanations of a rule are adopted by Laskey and Lehner (1989), by Lowrance et al (1986), and by Zarley et al (1988). This is also consistent with our explanation in section 2.2.

In section 5 we only discuss one situation involving the dependency problem. Hau and Kashyap (1990) discussed several situations based on their representation of heuristic rules.

### 6.2 SUMMARY

Evidential mappings are the main concept proposed in this paper. The extended Dempster-Shafer theory is more powerful for propagating beliefs and at the same time keeps all the features of the original theory. The following are main features in our approach: 1). representing uncertain relations between evidence spaces and hypothesis spaces by evidential mappings; 2). by creating evidential mappings for incomplete heuristic rule sets, more heuristic knowledge can be represented in D-S theory; 3). by constructing complete evidential mapping matrices any piece of evidence bearing on an evidence space can be propagated to the corresponding hypothesis space; 4). when a set of heuristic rules is detailed enough to form a Bayesian multi-valued causal link model, any result produced by Bayesian inference can also be carried out by D-S theory under evidential mappings; 5). evidential mappings are consistent with other previous research work in this respect; 6). a series of belief propagation procedures are easily deduced based on evidential mappings.

Heuristic knowledge is important in knowledge based systems. Representing this kind of knowledge and propagating beliefs are the main and the most difficult tasks for designers of knowledge based systems. This paper makes some progress in this area. Future work concerning evidential mappings in the Dempster-Shafer theory should focus on exploring more features of evidential mappings and more approaches to dealing with dependency relations.

## Acknowledgement

The research work was supported by 'Information Systems Committee of UFC' when the first author was working in the Dept. of Information Systems, University of Ulster.